# What-If Decision Support for Product Line Extension Using Conditional Deep Generative Models

**Yinxing Li[a] and Tsukasa Ishigaki[a*]**

[a] 27-1, Kawauchi, Aoba-ku, Sendai, Miyagi, 980-8576, Japan

Graduate School of Economics and Management, Tohoku University

e-mail: yinxing.li.a8@tohoku.ac.jp, isgk@tohoku.ac.jp

Corresponding Author: Tsukasa Ishigaki (isgk@tohoku.ac.jp)

**Abstract**

Product line extension is a strategically important managerial decision that requires anticipating how consumer segments and purchasing contexts may respond to hypothetical product designs that do not yet exist in the market. Such decisions are inherently uncertain because managers must infer future outcomes from historical purchase data without direct market observations. This study addresses this challenge by proposing a data-driven decision support framework that enables forward-looking what-if analysis based on historical transaction data. We introduce a Conditional Tabular Variational Autoencoder (CTVAE) that learns the conditional joint distribution of product attributes and consumer characteristics from large-scale tabular data. By conditioning the generative process on controllable design variables such as container type, volume, flavor, and calorie content, the proposed model generates synthetic consumer attribute distributions for hypothetical line-extended products. This enables systematic exploration of alternative design scenarios without costly market pretests. The framework is evaluated using home-scan panel data covering more than 20,000 consumers and 700 soft drink products. Empirical results show that the CTVAE outperforms existing tabular generative models in capturing conditional consumer attribute distributions. Simulation-based analyses further demonstrate that the generated synthetic data support knowledge-driven reasoning for assessing cannibalization risks and identifying potential target segments. These findings highlight the value of conditional deep generative models as core components of decision support systems for product line extension planning.

## 1. Introduction

Product line extension is a widely used managerial strategy for expanding product portfolios by introducing new variants under an existing brand name (Keller, 2008). By modifying product attributes such as size, flavor, packaging, or functionality, firms aim to enhance brand value, increase market coverage, and strengthen customer loyalty. At the same time, poorly designed extensions may dilute brand equity and increase operational and distribution costs, underscoring the importance of informed managerial decision-making in product line extension planning (Reddy et al., 1994).

A central challenge in such decisions is anticipating how consumer segments and purchasing contexts may respond to new product designs that have not yet been introduced to the market. Because these products are unobserved prior to launch, direct empirical evidence is unavailable, and managers must make decisions under substantial uncertainty. In practice, evaluations often rely on managerial experience, limited pretests, or post-launch sales analyses. Although marketing research has developed systematic approaches for identifying consumer needs and preferences (Crawford and Di Benedetto, 2010; Malhotra, 2019), pretest-based methods are typically costly, time-consuming, and constrained to evaluating a small number of design alternatives. As a result, they provide limited support for exploring the large and combinatorial design space associated with product line extension decisions.

From a decision support perspective, this problem can be framed as the task of transforming consumer-related data into actionable knowledge that enables forward-looking managerial reasoning. Decision-makers must evaluate hypothetical products and assess how changes in controllable product attributes may shift the distribution of consumer characteristics and usage contexts. For example, when extending a soft drink product line from a 500 mL container predominantly purchased by single consumers to a 2,000 mL container, managers must evaluate about whether the extension will attract new segments, such as families, or primarily cannibalize the existing customer base. Addressing this challenge requires computational decision support methods that go beyond descriptive analysis of historical consumer data and enable systematic what-if analysis for untested product designs.

One class of approaches that has sought to address such problems without relying on costly pretests is agent-based simulation. Agent-based models represent consumers as autonomous agents endowed with behavioral rules or cognitive characteristics and have been used to simulate competitive brand choice and market dynamics in artificial environments (Zhang and Zhang, 2007; Huiru et al., 2018). By examining interactions among heterogeneous agents, these models enable exploratory analysis of how variations in consumer attributes or environmental conditions may lead to emergent market-level outcomes. However, incorporating large-scale empirical purchase data into agent design and behavioral rule specification remains challenging. As a result, the correspondence between simulated outcomes and actual consumer behavior is often indirect and difficult to validate, limiting the reliability and managerial usefulness of the generated insights (Rand and Rust, 2011).

Recent advances in deep learning provide new opportunities to address these limitations by enabling computational models to acquire rich knowledge directly from large-scale empirical data. In the context of decision support for consumer-oriented problems, deep learning models are particularly attractive because they can capture nonlinear relationships and high-order interactions among multiple attributes. Among these approaches, deep generative models—such as variational autoencoders (VAEs) and generative adversarial networks (GANs)—offer a principled framework for data-driven knowledge generation. By learning the underlying data distribution, generative models can produce synthetic but plausible observations that represent potential outcomes beyond those directly observed in historical data. Although deep generative models have been extensively studied in unstructured domains such as images, audio, and video (Gm et al., 2020), their application to structured decision-making problems remains limited. Consumer behavior data are typically represented as heterogeneous tabular data that combine categorical and numerical variables and exhibit complex dependency structures. Accurately modeling these dependencies is essential for generating reliable knowledge that can support managerial decision-making. Existing applications of generative models for tabular data have primarily focused on data anonymization or augmentation in domains such as bioinformatics and healthcare (Sahakyan et al., 2021; Fonseca and Bacao, 2023), rather than on supporting forward-looking decisions.

Consumer behavior data are typically represented as tabular data that combine categorical and numerical variables and exhibit complex dependency structures. Accurately modeling these dependencies is essential for generating reliable knowledge that can support managerial decision-making. Existing applications of deep generative models for tabular data have primarily focused on data anonymization or augmentation in domains such as bioinformatics and healthcare (Sahakyan et al., 2021; Fonseca and Bacao, 2023), rather than on supporting forward-looking decision-making tasks. Learning conditional generative models from consumer purchase histories enables inference about how consumer attribute distributions may change under alternative, hypothetical product configurations. Such capabilities directly support what-if analysis, allowing managers to assess cannibalization risks, identify potential new segments, and evaluate trade-offs among competing design attributes without relying on costly pretests. However, leveraging deep generative models for strategic marketing decisions requires models that can handle tabular data and condition on controllable product design attributes.

To address these requirements, this study proposes a Conditional Tabular Variational Autoencoder (CTVAE) as a core component of a data-driven decision support framework for product line extension. The proposed model learns the conditional joint distribution of consumer characteristics given product attributes and generates synthetic consumer attribute distributions under specified design conditions. Product attributes such as container capacity, flavor, and other design variables are treated as controllable conditional inputs, while the outputs consist of interpretable consumer characteristics, including age, household composition, and income level. This conditional generative structure enables systematic what-if analysis by directly linking design choices to expected shifts in consumer segments.

Figure 1 presents an overview of the proposed decision support framework. The model is trained using large-scale home-scan panel data collected by Macromill, Inc., which record the purchase behaviors of 20,682 consumers over a one-year period in Japan and are statistically balanced to reflect national demographic distributions. The dataset comprises 206,561 purchase observations across 746 soft drink products. By learning from these empirical purchase histories, the proposed framework acquires data-driven knowledge that supports forward-looking inference about consumer attribute distributions under alternative product configurations. The generated knowledge provides actionable decision support for product line extension planning. In particular, it enables managers to assess potential cannibalization risks, identify consumer segments likely to be attracted by new product variants, and evaluate trade-offs among competing design attributes. Through these capabilities, the proposed CTVAE-based framework enhances the transparency and analytical rigor of product line extension decisions and illustrates how conditional deep generative models can be integrated into knowledge-based decision support systems for marketing and product design.

Fig. 1 Overall architecture of the proposed framework

## 2. Related Works

### 2.1 Product line extension and its support system

Product line extension has been extensively examined in the marketing and management literature, primarily as a form of brand extension. A large body of empirical research has focused on understanding how consumers' existing brand knowledge (Aaker and Keller, 1990) and previously formed attitudes (Boush and Loken, 1991) influence evaluations of extended products. Other studies have investigated moderating factors such as brand equity (Clark Sinapuelas and Sisodiya, 2010) and have analyzed product line extensions in specific contexts, including mathematical modeling (He et al., 2022), durable consumer goods (Park and Sela, 2020), and price competition (Kadiyali et al., 1998). While these studies have substantially advanced theoretical and empirical understanding of product line extension mechanisms and outcomes, their primary objective has been explanatory or retrospective, rather than to support managerial decision-making prior to market introduction.

From a decision support perspective, the key limitation of this stream of research lies in its limited ability to support forward-looking reasoning about hypothetical product designs. Most existing studies analyze observed extensions and realized market outcomes, offering valuable insights into why certain extensions succeed or fail, but providing limited operational support for evaluating untested design alternatives under uncertainty. As a result, managers often remain reliant on experiential judgment or costly pretests when making product line extension decisions.

Only a small number of studies have explicitly framed product line extension as a decision support or knowledge acquisition problem. Liao et al. (2008), for example, proposed a support system for product line expansion by constructing a relational database from purchase histories collected at Carrefour Taiwan stores and applying data mining techniques such as association rule mining and k-means clustering to extract customer knowledge. Their work demonstrated the potential value of leveraging historical transaction data for managerial support. However, the extracted knowledge was limited to patterns observed in existing products and did not enable inference about consumer behavior for unobserved or hypothetical product extensions.

More recently, deep learning techniques have been increasingly applied to consumer behavior and marketing-related problems. Prior studies have focused on tasks such as extracting consumer characteristics (Ładyżyński et al., 2019; Sun et al., 2021), predicting consumer behavior (Mirashk et al., 2019; Zhu et al., 2023; Liu et al., 2024; Mamta and Sangwan, 2024; Zhang et al., 2024), designing collaborative or pricing strategies (Carlo et al., 2021), and developing generative AI–based chatbots for marketing applications (Chan and Choi, 2025). Although these approaches demonstrate the analytical power of deep learning, they predominantly adopt a predictive or descriptive orientation and do not aim to support what-if analysis of product design decisions.

However, these approaches predominantly adopt a predictive or analytical perspective and do not aim to generate conditional knowledge about how consumer attribute distributions may change in response to hypothetical product design decisions. Consequently, the research objectives of existing studies differ fundamentally from those of the present work, which focuses on generative knowledge acquisition and inference to support strategic decision-making under uncertainty. In contrast, decision support for product line extension requires generative inference capabilities that enable managers to reason about how consumer attribute distributions may change in response to alternative and as-yet-unobserved product designs. Existing studies rarely address this requirement explicitly, leaving a gap between advances in consumer analytics and the needs of strategic decision-making under uncertainty. The present study addresses this gap by focusing on conditional knowledge generation and inference as a foundation for data-driven decision support in product line extension planning.

### 2.2 Synthetic data generator for tabular data

Deep learning methods for tabular data analysis have been actively studied and have produced substantial results in various application domains (Sahakyan et al., 2021; Fonseca and Bação, 2023; Borisov et al., 2024). However, the problem of generating high-quality synthetic tabular data has received comparatively less attention, particularly from the perspective of knowledge acquisition and inference (Borisov et al., 2024). Early approaches to tabular data generation, including medWGAN (Choi et al., 2017), Cramér GAN (Mottini et al., 2018), and TableGAN (Park et al., 2018), were primarily motivated by privacy preservation and data anonymization, especially in medical and healthcare contexts (Choi et al., 2017; Nikolentzos et al., 2023). While these models demonstrated that deep generative techniques could capture complex dependencies in tabular data, they were not designed to support inference about hypothetical scenarios or to generate knowledge tailored to specific decision variables.

Constructing effective generative models for tabular data presents challenges that differ fundamentally from those encountered in image or text domains. Two issues are particularly critical. First, tabular data typically consist of heterogeneous variables, requiring the simultaneous modeling of continuous and discrete attributes with fundamentally different statistical properties. Second, categorical variables often exhibit highly imbalanced frequency distributions, which can hinder the learning of rare categories when standard mini-batch training strategies are applied. To address these challenges, Xu et al. (2019) proposed two influential generative models for tabular data: Conditional Tabular Generative Adversarial Networks (CTGAN) and Tabular Variational Autoencoders (TVAE). These models introduce mode-specific normalization to transform non-Gaussian continuous variables into mixtures of Gaussian distributions, enabling unified input representations for heterogeneous attributes. In addition, they employ conditional training strategies that allow the models to learn effectively from imbalanced categorical data by conditioning the learning process on discrete variable categories.

However, in CTGAN and TVAE, conditional distributions are primarily utilized as mechanisms for stabilizing the training process, rather than as explicit tools for conditional data generation and inference. As a result, these models do not directly support the generation of synthetic samples from user-specified conditional distributions. Similarly, other tabular generative models, such as VAEM (Ma et al., 2020), artGAN (Fan et al., 2020), and TAEI (Darabi and Elor, 2021), are not designed to generate data conditioned on explicit control variables. While conditional generative models such as conditional VAEs have been extensively studied (Kingma et al., 2014; Sohn et al., 2015; Pandey and Dukkipati, 2017; Tang et al., 2023), these approaches do not target tabular data representations.

Bayesian network-based approaches have also been explored for synthetic data generation (Zhang et al., 2017); however, due to their algorithmic structure, they are not well suited for generating samples from arbitrary conditional distributions in high-dimensional tabular settings. In contrast, the model proposed in this study explicitly generates synthetic tabular data from conditional distributions, enabling controlled knowledge generation and inference. This distinction differentiates the proposed CTVAE from existing tabular generative models such as CTGAN and TVAE, both in terms of model structure and intended use within knowledge-based decision support systems.

## 3. Conditioning Tabular VAE

### 3.1 Tabular data generation with conditional input

A property of tabular data is that the relationship between variables follows a joint probability distribution that is in many cases unknown. In this section, we consider the problem of generating synthetic data that follows a joint distribution of other variables when conditioning some variables in tabular data.

Consider tabular data $T$ of size $N \times M$. Each row $i$ of $T$ contains the sample $i$, and each column $j$ contains the $j$-th random variable. The typical objective of synthetic tabular data generation is to make a generator $G$ that samples from a target joint distribution by learning of $T$. Here, we assume that the random variables in each column of $T$ can be partitioned into variables $x_{s,j}$ $(j = 1, \cdots, M_s)$ that should be generated as synthetic data and variables $x_{c,j}$ $(j = 1, \cdots, M_c)$ that are the conditional parts controlling the generation of synthetic data ($M_s + M_c = M$). The elements in row $i$ and column $j$ of $T$ are denoted as $x_{s,j,i}$ or $x_{c,j,i}$. For application involving product-line extensions, the variables $x_{s,j}$ and $x_{c,j}$ represent the customer and product attributes, respectively. While typical tabular data generation focuses on joint distributions $p(x_{s,1}, \cdots, x_{s,M_s})$, the proposed method constructs a generator that models the conditional joint distribution $p(x_{s,1}, \cdots, x_{s,M_s} | x_{c,1}, \cdots, x_{c,M_c})$.

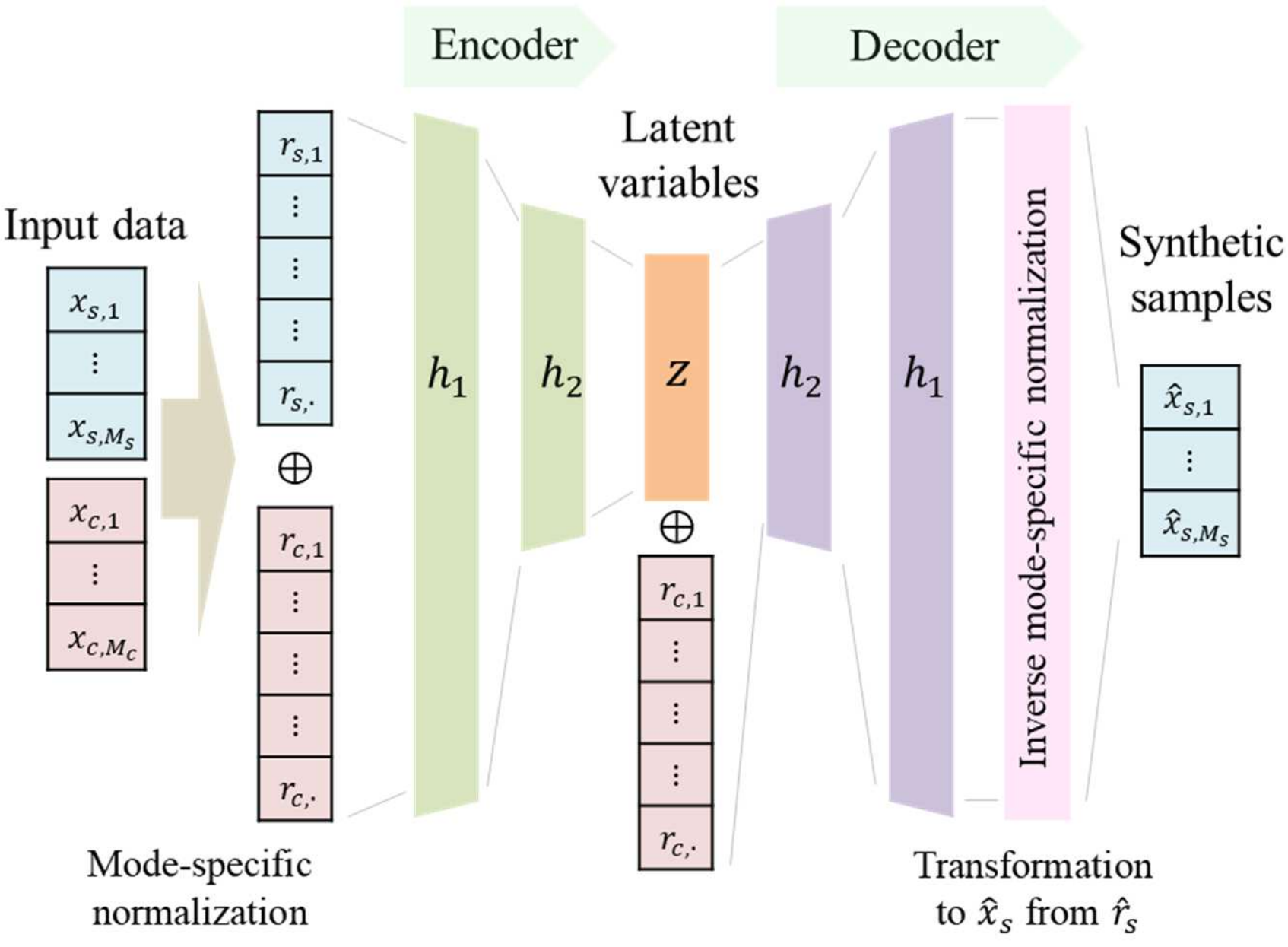

Figure 2. Architecture of the proposed model

### 3.2 Proposed model

Figure 2 illustrates the architecture of the proposed conditional tabular VAE (CTVAE). Here, the input vectors are written as $\boldsymbol{x}_s = \left[x_{s,1}, \cdots, x_{s,M_s}\right]^T$ and $\boldsymbol{x}_c = \left[x_{c,1}, \cdots, x_{c,M_c}\right]^T$. Similar to the VAE and TVAE, the CTVAE consists of an encoder with input data $\boldsymbol{x}_s$ and $\boldsymbol{x}_c$ and a decoder with latent variable $\boldsymbol{z}$ as input. Here, the encoder's probability distribution in CTVAE is expressed as $q_\phi(\boldsymbol{z}|\boldsymbol{x}_s, \boldsymbol{x}_c)$ and that of the decoder as $p_\theta(\boldsymbol{x}_s|\boldsymbol{z}, \boldsymbol{x}_c)$. For $\boldsymbol{x}_s$ and $\boldsymbol{x}_c$ with tabular data characteristics, we perform mode-specific normalization, as proposed by Xu et al. (2019). Mode-specific normalization transforms each variable into tabular data with continuous and discrete variables, both as appropriate inputs to a deep neural network. For continuous variables, the shape of the distribution is modeled using a mixture of Gaussian distributions, and normalization was performed for each component of each Gaussian distribution. The number of mixtures in the Gaussian distribution is estimated using the input data. Here, the mode-specific normalized $\boldsymbol{x}_s$ and $\boldsymbol{x}_c$ are denoted as $\boldsymbol{r}_s$ and $\boldsymbol{r}_c$, respectively. The CTVAE was modeled using an end-to-end deep neural network, and $\boldsymbol{r}_s$ and $\boldsymbol{r}_c$ are given as inputs. Because each column of the input data is not necessarily independent of the others, we employ a fully connected network structure in the CTVAE. We modeled the CTVAE as follows:

1. Input the vector $\boldsymbol{r}_s \oplus \boldsymbol{r}_c$

2. Encoder
$$\begin{cases} \boldsymbol{h}_1 = \mathrm{ReLU}\big(\mathrm{AFFINE}(\boldsymbol{r}_s \oplus \boldsymbol{r}_c)\big), (\mathit{dimension}\text{: } |\boldsymbol{r}_s \oplus \boldsymbol{r}_c| \to 256) \\ \boldsymbol{h}_2 = \mathrm{ReLU}\big(\mathrm{AFFINE}(\boldsymbol{h}_1)\big), (\mathit{dimension}\text{: } 256 \to 128) \\ \boldsymbol{\mu} = \mathrm{AFFINE}(\boldsymbol{h}_2), (\mathit{dimension}\text{: } 128 \to 128) \\ \boldsymbol{\sigma}^2 = \exp\{0.5 \times \mathrm{AFFINE}(\boldsymbol{h}_2)\}, (\mathit{dimension}\text{: } 128 \to 128) \\ \boldsymbol{z} \sim N(\boldsymbol{\mu}, \boldsymbol{\sigma}^2 I) \end{cases}$$

3. Decoder
$$\begin{cases} \boldsymbol{h}_1 = \mathrm{ReLU}\big(\mathrm{Affine}(\boldsymbol{z} \oplus \boldsymbol{r}_c)\big), (\textit{dimension}\colon |\boldsymbol{z} \oplus \boldsymbol{r}_c| \to 128) \\ \boldsymbol{h}_2 = \mathrm{ReLU}\big(\mathrm{Affine}(\boldsymbol{h}_1)\big), (\textit{dimension}\colon 128 \to 256) \\ p_\theta(\boldsymbol{r}_s|\boldsymbol{z}, \boldsymbol{r}_c) = \mathrm{TVAE}(\boldsymbol{h}_2) \end{cases}$$

, where ReLU is the ReLU activation function, AFFINE is the Affine join function for all joins, $\oplus$ is the function to create a concatenate vector, and TVAE is the sample generation function for the categorical distribution of the decoder part of TVAE as proposed in Xu et al. (2019). The description of *dimension* in the above equation is the number of dimensions that showed the best performance as a result of the verification described below.

The estimated generator by tabular data $T$ as training data is written as $\hat{G} \equiv \{\hat{q}_\phi(\boldsymbol{z}|\boldsymbol{r}), \hat{p}_\theta(\boldsymbol{r}_s|\boldsymbol{z}, \boldsymbol{r}_c)|T\}$. Then, each synthetic sample $i$ conditioned by variables $\boldsymbol{x}_c$ can be generated as $\hat{\boldsymbol{x}}_{s,i} \sim \mathrm{i.i.d.}\, \hat{G}(\boldsymbol{x}_s|\boldsymbol{x}_c)$, $\left(\hat{\boldsymbol{x}}_{s,i} = \left[\hat{x}_{s,j=1,i}, \cdots, \hat{x}_{s,j=M_s,i}\right]^T\right)$.

Consider maximizing the conditional log-likelihood $\log\{p(\boldsymbol{x}_s|\boldsymbol{x}_c)\}$ of the CTVAE decoder. Let the variational lower bound (ELBO) be $L(\boldsymbol{x}_s, \boldsymbol{z}|\boldsymbol{x}_c)$,

$$\log p_\theta(\boldsymbol{x}_s|\boldsymbol{x}_c) - L(\boldsymbol{x}_s, \boldsymbol{z}|\boldsymbol{x}_c) = \log p_\theta(\boldsymbol{x}_s|\boldsymbol{x}_c) - \int q_\phi(\boldsymbol{z}|\boldsymbol{x}_s, \boldsymbol{x}_c) \log \frac{p_\theta(\boldsymbol{x}_s, \boldsymbol{z}|\boldsymbol{x}_c)}{q_\phi(\boldsymbol{z}|\boldsymbol{x}_s, \boldsymbol{x}_c)} d\boldsymbol{z}$$

$$= \int q_\phi(\boldsymbol{z}|\boldsymbol{x}_s, \boldsymbol{x}_c) \log p_\theta(\boldsymbol{x}_s|\boldsymbol{x}_c)\, d\boldsymbol{z} - \int q_\phi(\boldsymbol{z}|\boldsymbol{x}_s, \boldsymbol{x}_c) \log \frac{p_\theta(\boldsymbol{z}|\boldsymbol{x}_s, \boldsymbol{x}_c) p_\theta(\boldsymbol{x}_s|\boldsymbol{x}_c)}{q_\phi(\boldsymbol{z}|\boldsymbol{x}_s, \boldsymbol{x}_c)} d\boldsymbol{z}$$

$$= \int q_\phi(\boldsymbol{z}|\boldsymbol{x}_s, \boldsymbol{x}_c) \log \frac{q_\phi(\boldsymbol{z}|\boldsymbol{x}_s, \boldsymbol{x}_c)}{p_\theta(\boldsymbol{z}|\boldsymbol{x}_s, \boldsymbol{x}_c)} d\boldsymbol{z} = KL\big\{q_\phi(z|\boldsymbol{x}_s, \boldsymbol{x}_c), p_\theta(z|\boldsymbol{x}_s, \boldsymbol{x}_c)\big\}.$$

The maximization of the conditional log-likelihood $\log p_\theta(\boldsymbol{x}_s|\boldsymbol{x}_c)$ of CTVAE is achieved by minimizing $KL\big\{q_\phi(z|\boldsymbol{x}_s, \boldsymbol{x}_c), p_\theta(z|\boldsymbol{x}_s, \boldsymbol{x}_c)\big\}$, where

$$L(\boldsymbol{x}_s, \boldsymbol{z}|\boldsymbol{x}_c) = -KL\big\{q_\phi(z|\boldsymbol{x}_s, \boldsymbol{x}_c), p_\theta(z|\boldsymbol{x}_c)\big\} + \int q_\phi(z|\boldsymbol{x}_s, \boldsymbol{x}_c) \log p_\theta(\boldsymbol{x}_s|\boldsymbol{z}, \boldsymbol{x}_c)\, d\boldsymbol{z}$$

In VAE, $p_\theta(z) = N(\boldsymbol{0}, I)$ is assumed, while in CTVAE, $p_\theta(z|\boldsymbol{x}_c) = N(\boldsymbol{0}, I)$ is assumed for learning.

## 4. Synthetic Data Generation

### 4.1 Dataset

#### 4.1.1 Consumer’s purchase history and attributes

We constructed a generator $G$ using QPR data from a database of consumer purchase history owned by Macromill, Inc. Daily purchase histories were recorded by each of the approximately 30,000 monitors using an in-house barcode reading system. The following information on the monitors is recorded: prefecture of residence in Japan (47 categorical variables), age (continuous data), gender (2 categorical variables), marital status (3 categorical variables), presence of children (2 categorical variables), occupation (13 categorical variables), family structure (5 categorical variables), housing type (6 categorical variables), household income (14 categorical variables), purchase quantity (continuous variable), product user (3 categorical variables), purchase time (6 categorical variables), and purchase season (4 categorical variables). In this experiment, variable vector $\boldsymbol{x}_s$ consists of these consumer attributes, which are the target variables to be synthesized from generator $\hat{G}(\boldsymbol{x}_s|\boldsymbol{x}_c)$.

#### 4.1.2 Product attributes

Generator $\hat{G}(\boldsymbol{x}_s|\boldsymbol{x}_c)$ learned from the data of 746 soft drinks that were purchased frequently, and for which product attribute data could be identified. The attribute data for each product were assigned by referring to each brand's website and other sources. Here, the product name, manufacturer name, country of origin, container type (can, plastic bottle, etc.), content volume (ml), calories (kcal / 100 ml), and ingredient names were assigned as product attributes. There were 312 unique names for the ingredients, including lemon juice, carbonation, flavoring, citric acid, etc. For each product, the top five listed ingredient names were assigned as attributes. For products with fewer than five listed ingredient names, "none" was assigned as a product attribute until there were five ingredient types. In our experiments, we employ these product attributes as $\boldsymbol{x}_c$ to generate synthetic samples $\hat{\boldsymbol{x}}_{s,i}$ from $\hat{G}(\boldsymbol{x}_s|\boldsymbol{x}_c)$.

In the experiment, 206,561 purchase histories of 20,682 consumers with at least one purchase history of 748 soft drinks during the one-year period from October 1, 2018, to September 30, 2019, were used to learn the generator $\hat{G}(\boldsymbol{x}_s|\boldsymbol{x}_c)$. $\boldsymbol{x}_s$ consists of 13 variables and $\boldsymbol{x}_c$ consists of 11 variables. The dimension of variable $\boldsymbol{r}_s$ is 140, and the dimension of $\boldsymbol{r}_c$ is 1,273 with mode-specific normalization. Therefore, the input vector to the encoder of the CTVAE was 1,413-dimensional vectors.

### 4.2 Experimental results

#### 4.2.1 Evaluation metrics

To tune the dimensions of the intermediate layer of the CTVAE, we evaluate the performance of the generator with the KS complement based on Kolmogorov-Smirnov statistics for continuous variables and the TV complement for discrete variables (SDMetrics DataCebo (a) 2023 and SDMetrics DataCebo (b) 2023). The indicators are quantitative measures of the difference between the distribution of each stochastic variable $x_{s,j}$ and the distribution of the synthetic data $\hat{x}_{s,j}$, and have been employed by Titar and Ramanathan (2024) and Johann et al., (2025). Given a data set $\{\boldsymbol{x}_{s,j}\} \equiv \{x_{s,j,i=1}, \cdots, x_{s,j,i=N}\}$ for a continuous variable $j$, let $F(\{\boldsymbol{x}_{s,j}\})$ be the function that returns its empirical cumulative distribution function, $\|\cdot\|_\infty$ be the upper bound ($H_\infty$ norm) for the entire continuous variable domain, and $\mathrm{hist}(\{\boldsymbol{x}_{s,j}\})$ be the function that returns its normalized histogram for a discrete variable $j$. Then, the KS complement and the TV complement are expressed by the following equations, respectively,

$$KS_j = 1 - \left\|F(\{\boldsymbol{x}_{s,j}\}) - F(\{\hat{\boldsymbol{x}}_{s,j}\})\right\|_\infty,$$

$$TV_j = 1 - \frac{1}{2}\sum \left|\text{hist}(\{\boldsymbol{x}_{s,j}\}) - \text{hist}(\{\hat{\boldsymbol{x}}_{s,j}\})\right|.$$

The measures are $0 \leq KS_j, TV_j \leq 1$, and the value is close to 1 when the distribution of the test data is close to the distribution of the synthetic data. Here, the $KS_{p,j}$ and $TV_{p,j}$ of each variable $j$ and product $p$ were calculated, and the mean complement is calculated as follows:

$$MC_p = \frac{1}{M_s}\left(\sum_{j \in CV} KS_{p,j} + \sum_{j \in DV} TV_{p,j}\right),$$

where $CV$ represents a set of continuous variables and $DV$ represents a set of discrete variables. Similarly, we also used the weighted metric $wMC_p$ where $MC_p$ is adjusted by the number of purchases of the product in the test data ($|I_p|$), to evaluate the prediction performance.

$$wMC_p = \frac{|I_p|}{\sum_{k=1}^{M_s} |I_k|}\left(\sum_{j \in CV} KS_{p,j} + \sum_{j \in DV} TV_{p,j}\right).$$

Table 1 Descriptive statistics of training and test data

| | Number of product | Average of purchase | Average of variance of purchase | Minimum number of purchases | Maximum number of purchases |
|---|---|---|---|---|---|
| Training data | 674 | 274.28 | 582.80 | 1 | 6,396 |
| Test data | 72 | 301.30 | 583.84 | 4 | 2,972 |

#### 4.2.2 Prediction performance using hold-out samples

To validate the prediction performance, 674 products were randomly assigned as training data and 72 products were assigned as test data. Because data on approximately 600 products were required for suitable learning by the generator, the training and test data were split into this proportion. Table 1 presents the descriptive statistics of the training and test data. Generator $\hat{G}$ was trained using the 674 products assigned to the training data. In the learning process, 10% of the training data was used as the validation dataset to optimize the number of epochs by validation loss. Let $\boldsymbol{x}_{c,p}^{test}$ be vectors of the variables of product $p$ assigned to the test data. For each of the 72 products assigned to the test data, we generated synthetic data for sample $i$ of product $p$ as $\hat{\boldsymbol{x}}_{s,i}^{(p)} \sim \text{i.i.d.}\ \hat{G}(\boldsymbol{x}_{s,j}|\boldsymbol{x}_{c,p}^{test})$ and measured the prediction performance of generator $\hat{G}$ on unknown products using $MC_p$ and $wMC_p$.

The number of each intermediate layer's dimensions is verified using the following settings: 64 dimensions (64–32–32–32–64), 128 dimensions (128–64–64–64–128), 256 dimensions (256–128–128–128–256), and 512 dimensions (512–256–256–256–512). The numbers in parentheses are the

number of dimensions in the encoder's first and second layers, the number of dimensions of latent space and the number of dimensions in the first and second layers of the decoder, respectively.

To the best of our knowledge, no synthetic data generator for tabular data exists that can efficiently perform conditional sampling on high-dimensional conditional variables, which is the objective here. We therefore use synthetic samples from the CTGAN and TVAE as a baseline to compare the performances of the tabular data generators. However, conditional sampling from the CTGAN and TVAE incurs very high computational costs, making the sampling method impractical. Note that the evaluation results for the CTGAN and TVAE are not sampled from the conditional distribution, but from the joint distribution $\hat{G}(\boldsymbol{x}_s, \boldsymbol{x}_c)$. In contrast, the results of CTVAE are based on sampling from $\hat{G}(\boldsymbol{x}_s|\boldsymbol{x}_c)$.

Table 2 lists the average values of $MC_p$ and $wMC_p$ of the products assigned to the test data for CTGAN, TVAE, and CTVAE. The results are calculated for 30,000 samples for each product.

Table 2. Prediction performance of TVAE, CTGAN and CTVAE

| | Average of MC | | | |
|---|---|---|---|---|
| | 64 dim | 128 dim | 256 dim | 512 dim |
| TVAE | 0.704 | 0.729 | 0.737 | 0.736 |
| CTGAN | 0.742 | 0.739 | 0.745 | 0.734 |
| CTVAE | 0.745 | 0.758 | **0.764** | 0.761 |
| | Weighted average of MC | | | |
| TVAE | 0.772 | 0.809 | 0.833 | 0.824 |
| CTGAN | 0.832 | 0.831 | 0.831 | 0.833 |
| CTVAE | 0.835 | 0.856 | **0.866** | 0.865 |

**4.2.3 Discussion for prediction performance**

The results summarized in Table 2 provide several insights into the predictive and inferential capabilities of the proposed CTVAE.

First, the CTVAE consistently outperforms CTGAN and TVAE in terms of predictive accuracy. Although these models differ in their sampling and training mechanisms, the superior performance of the CTVAE indicates that it acquires more faithful representations of the underlying relationships between product attributes and consumer characteristics. This finding suggests that the proposed model functions as a more reliable generative knowledge model, rather than merely achieving incremental improvements in prediction metrics. Second, for both MC averaging and weighted MC averaging, variations in the dimensionality of the intermediate latent layer did not lead to substantial differences in predictive performance. Among the evaluated configurations, the 256-dimensional latent representation achieved the best average performance across evaluation measures.

This result indicates that the proposed framework is relatively robust to changes in model capacity within a reasonable range, and that a moderate latent dimensionality is sufficient to capture the relevant dependency structures in the data. Accordingly, subsequent analyses are based on the 256-dimensional CTVAE configuration. Third, across all experimental settings, weighted MC averaging consistently outperformed simple MC averaging. This result highlights the importance of incorporating information from a larger number of observed purchase histories when generating synthetic data.

Overall, these results demonstrate that the proposed CTVAE not only achieves superior predictive performance but also provides a robust and effective mechanism for generating high-quality knowledge suitable for inference and decision support in structured consumer behavior modeling tasks.

### 4.3 Validations of synthetic data by examples

#### 4.3.1 An example of line extension for container and calorie

This subsection presents a case-based validation of the proposed framework from a decision support perspective, illustrating how the generated synthetic data can be used to reason about hypothetical product line extension scenarios. The example focuses on two soft drink products, A1 and A2, marketed under the same brand. Although the product contents are identical, the two products differ in key controllable design attributes: Product A1 (0 kcal per 100 mL) is sold in a 500 mL plastic bottle, whereas Product A2 (33 kcal per 100 mL) is sold in a 350 mL aluminum can. This setting provides a suitable decision context for examining whether the proposed model can support managerial inference about how changes in container type, volume, and calorie content influence consumer purchasing segments.

To validate the decision support capability of the proposed approach, we compare observed purchasing patterns with synthetic consumer attribute distributions generated under counterfactual product design conditions. Specifically, we analyze differences in purchasing segments defined by household composition (households with children versus households without children). Figure 3 summarizes the results of this comparative analysis and illustrates how the framework supports what-if reasoning for product line extension decisions.

The notation "A2 → A1" denotes a counterfactual scenario in which synthetic consumer data are generated by conditioning on the observed attributes of Product A2 while modifying the container, volume, and calorie attributes to match those of Product A1. Conversely, "A1 → A2" denotes the scenario in which synthetic data are generated by conditioning on Product A1 while modifying these attributes to match those of Product A2. Formally, the synthetic samples for the "A2 → A1" scenario are generated according to

$$\hat{G}(\boldsymbol{x}_s | \text{Volume} = 500\text{ mL}, \text{Container} = \text{Plastic bottle}, \text{Calories} = 0, \text{Others} = \text{Product A2}),$$

where "Others = Product A2" indicates that all conditional attributes other than volume, container, and calorie content—such as product name, manufacturer, country of origin, and ingredient information— are fixed to the observed attributes of Product A2. Similarly, the "A1 → A2" scenario is defined as

$$\hat{G}(\boldsymbol{x}_s|\text{Volume} = 350\text{ mL}, \text{Container} = \text{Can}, \text{Calories} = 33, \text{Others} = \text{Product A1}).$$

The scan panel dataset contains 887 observed purchase records for Product A1 and 409 for Product A2. To ensure stable estimation of consumer attribute distributions, 30,000 synthetic samples were generated for each conditional scenario. In the observed data, Product A1 exhibits a purchase ratio of 0.59 for households with children and 0.41 for households without children, whereas Product A2 shows a ratio of 0.47 and 0.53, respectively.

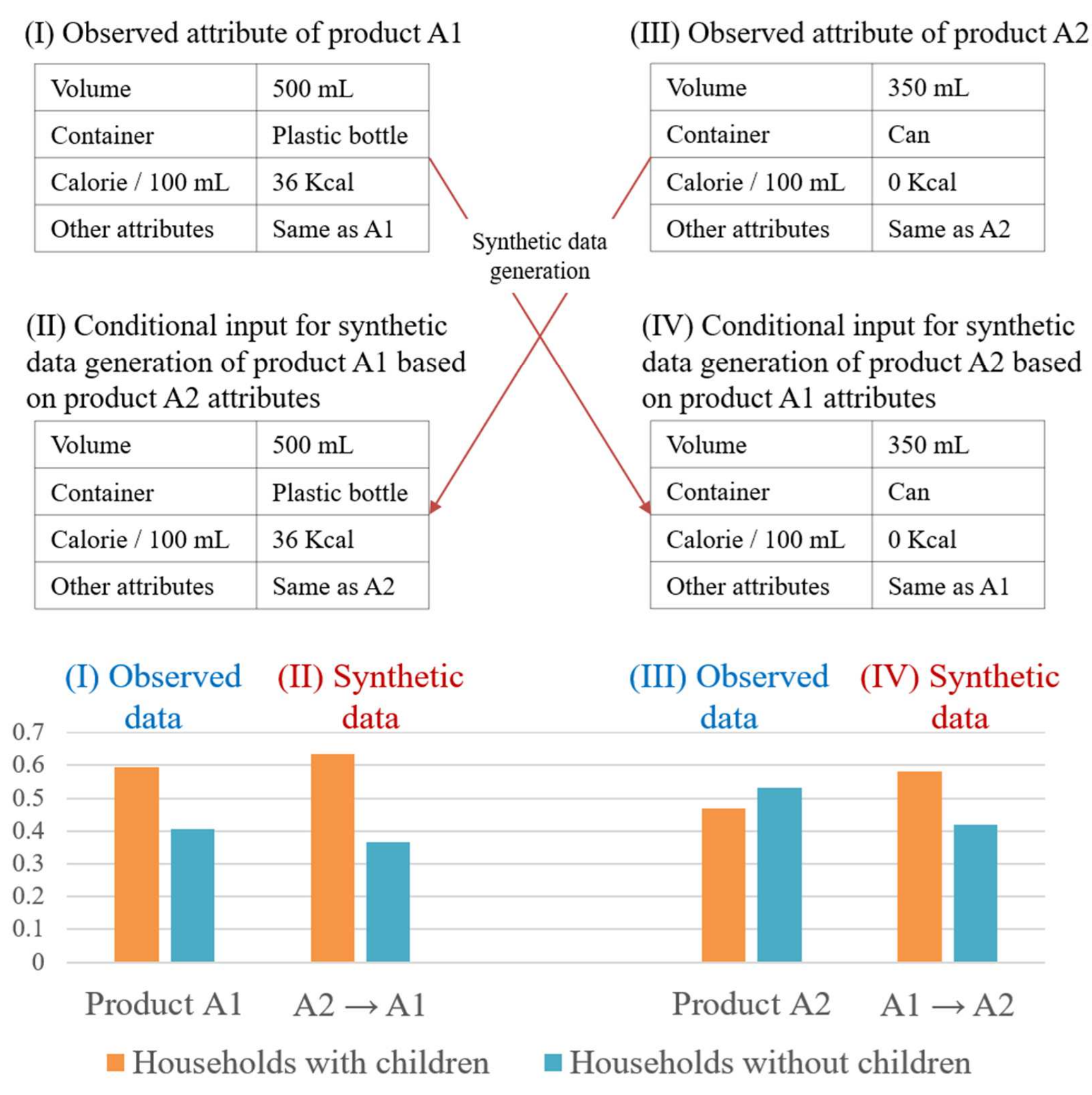

Figure 3. A validation results of synthetic data on volumes and containers for households with or without children in product A1 and A2

The synthetic results exhibit systematic shifts consistent with changes in product attributes. In the "A2 → A1" scenario, the generated data indicate a purchase ratio of 0.63 for households with children and 0.37 for households without children, reflecting a shift toward family-oriented segments. In contrast, the "A1 → A2" scenario yields a ratio of 0.58 and 0.42, respectively, moving the distribution toward that observed for Product A2, although the shift is not complete. Notably, while the observed data for Product A2 indicate a higher purchase share among households without children, this relationship is reversed in the counterfactual "A2 → A1" scenario, where households with children emerge as the dominant segment.

From a decision support perspective, these results demonstrate that the proposed CTVAE generates synthetic consumer data that respond meaningfully and directionally to changes in controllable product design attributes. The example illustrates how the framework enables counterfactual reasoning about unobserved product variants and supports managerial assessment of potential segment shifts prior to market introduction. This validation underscores the usefulness of the proposed approach as a knowledge acquisition and inference mechanism for analyzing product line extension decisions under uncertainty.

**4.3.2 An example of line extension for flavor**

This subsection provides a decision-oriented validation of the proposed framework in a flavor-based product line extension scenario. The analysis focuses on two soft drink products, B1 and B2, marketed under Brand B. While the two products share the same brand identity, they differ in flavor-related and ingredient attributes. Product B1 is a non-sugar, calorie-free product with 2,339 observed purchase records, whereas Product B2 is a lemon-flavored product with 1,079 observed purchase records. This setting offers an appropriate decision context for examining whether the proposed model can support managerial inference about how changes in flavor and ingredient attributes influence purchasing situations.

From a decision support perspective, the analysis evaluates whether the proposed framework can generate meaningful knowledge about situational shifts associated with alternative flavor designs. In this example, the situational attribute of interest is the season in which purchases occur, which is a relevant contextual factor for marketing decisions. Figure 4 summarizes the observed and synthetic distributions of purchase season for the two products, noting that the configuration of the horizontal axis differs from that used in Figure 3.

Four distributions are compared: (i) the observed purchase data for Product B1, (ii) the observed purchase data for Product B2, (iii) synthetic data generated by conditioning on the attributes of Product B2 while modifying ingredient-related attributes to match those of Product B1, and (iv) synthetic data generated by conditioning on the attributes of Product B1 while modifying ingredient-related attributes to match those of Product B2. The notation "B2 → B1" denotes a counterfactual scenario in which synthetic samples are generated according to

$$\hat{G}(\boldsymbol{x}_s|\mathrm{I1}=\mathrm{Fiber},\mathrm{I2}=\mathrm{Sweetener},\mathrm{Calorie}=0,\mathrm{Others}=\mathrm{Product\ B2}),$$

where $I_1$ and $I_2$ represent the first and second ingredient attributes, respectively. Conversely, "B1 → B2" denotes the scenario in which synthetic samples are generated as

$$\hat{G}(\boldsymbol{x}_s|\mathrm{I1}=\mathrm{Sugar},\mathrm{I2}=\mathrm{Potassium\ citrate},\mathrm{Calorie}=38,\ \mathrm{Others}=\mathrm{Product\ B1}).$$

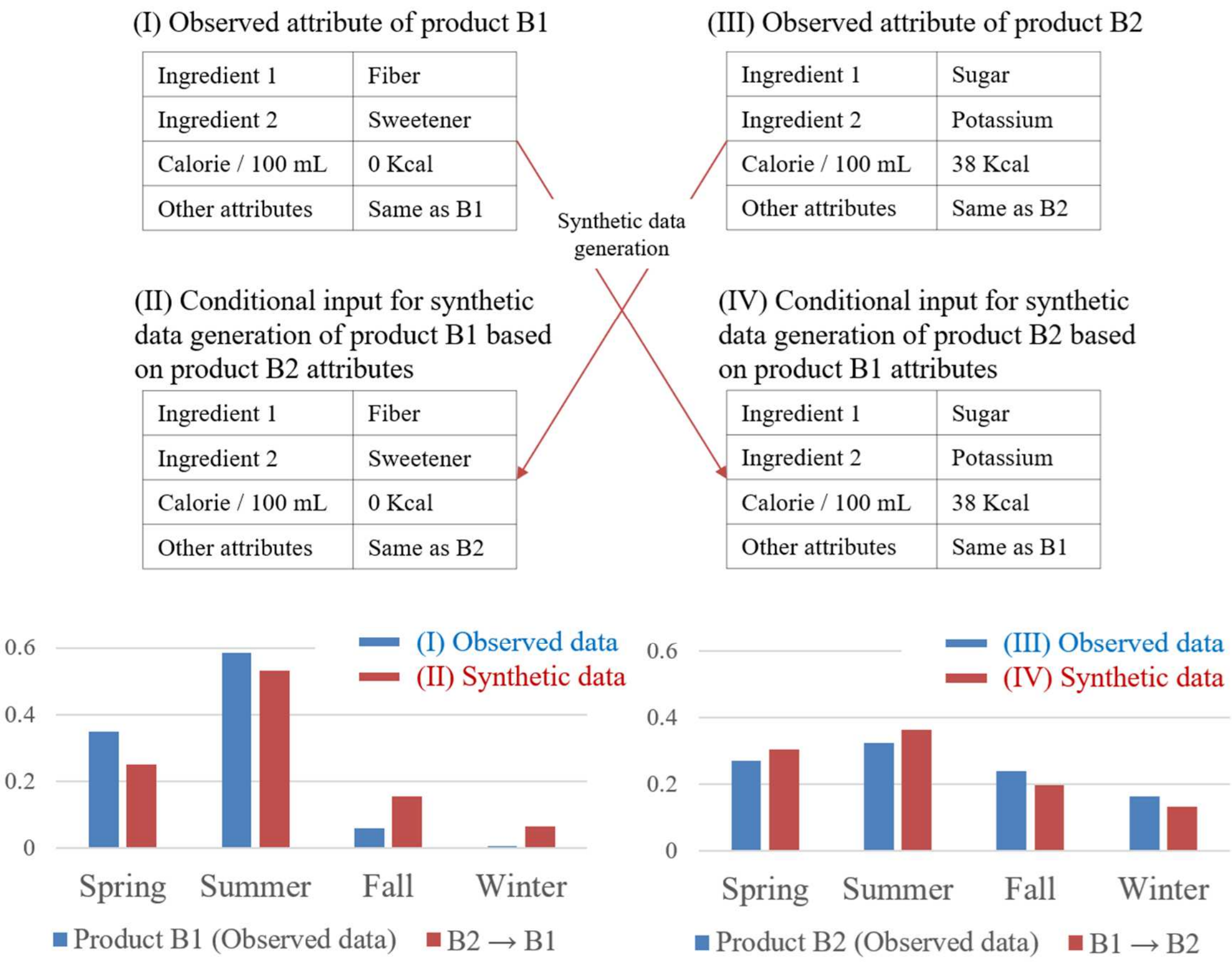

Figure 4. A validation results of synthetic data on ingredients for purchased season with product B1 and B2

As shown in Figure 4, both the observed and synthetic data exhibit coherent and comparable distributional patterns. The observed data for Product B1 show a pronounced seasonal trend, with higher purchase ratios during summer and lower ratios during fall and winter, suggesting usage in specific consumption contexts. In contrast, Product B2 exhibits relatively stable purchasing patterns across seasons, indicating weaker seasonal dependence. Importantly, the synthetic data generated under the corresponding conditional scenarios reproduce these tendencies, demonstrating alignment between observed behavior and inferred outcomes under hypothetical flavor configurations.

Together with the container and calorie example presented in Section 4.3.1, these results provide convergent evidence that the proposed CTVAE generates synthetic consumer data that respond systematically to changes in product attributes. From a decision support standpoint, this example illustrates how the framework enables conditional reasoning about situational purchasing contexts and supports knowledge validation for flavor-based product line extension decisions. The findings further demonstrate the potential of the proposed approach to assist managers in evaluating how alternative flavor designs may alter consumption situations prior to market introduction.

## 5. Discussion

### 5.1 Implications from CTVAE for practical marketing

The proposed CTVAE provides a data-driven decision support mechanism for exploring the implications of product line extensions under multiple hypothetical design scenarios. Rather than relying solely on post-hoc sales analysis or costly test marketing, the framework enables decision-makers to conduct forward-looking inference by simulating counterfactual product configurations and examining the resulting shifts in consumer attribute distributions. From a decision support systems perspective, this capability facilitates structured reasoning under uncertainty and complements traditional marketing research methods.

As a first illustrative decision scenario, we consider a line extension in which Product C, with 3,267 observed purchase records, is reformulated by changing its container from a plastic bottle to a pouch. Figure 5 compares the observed purchase-age distribution of Product C with the synthetic distribution generated by conditioning on the pouch container. The inferred results indicate an increase in the proportion of consumers in their 60s and a decrease in those in their 40s, while the purchase shares of consumers under 20, in their 20s, 50s, and 70s or older remain largely unchanged. From a managerial decision-making perspective, this pattern suggests that container attributes encode latent information about consumer preferences that can be externalized through conditional generative modeling. The inferred knowledge implies that a pouch-based line extension may benefit from targeted marketing strategies aimed at older consumers, particularly those in their 60s.

A second decision scenario examines a flavor-based line extension for Product D, a zero-calorie carbonated beverage with 708 observed purchase records, by introducing an apple cider vinegar flavor. Figure 6 presents the observed and synthetic distributions of purchase purposes for Product D. The synthetic results indicate an approximate 11 percentage point decrease in purchases for self-consumption and a corresponding increase of about 13 percentage points in purchases intended for family members or friends. This inferred shift suggests that the added flavor attribute alters the situational context in which the product is consumed. From a decision support standpoint, the model captures how flavor-related design choices influence consumption occasions, thereby generating actionable knowledge for designing product positioning, packaging messages, and communication strategies aligned with social or shared consumption contexts.

Taken together, these examples demonstrate how the proposed CTVAE can be operationalized as a practical knowledge-based decision support system for marketing. By generating synthetic consumer data conditioned on hypothetical product attributes, the framework enables marketers to perform systematic what-if analyses and to extract interpretable, decision-relevant knowledge without conducting preliminary market experiments. Because the approach is simulation-based, a broad range of alternative design and marketing scenarios can be evaluated efficiently. Consequently, the proposed CTVAE has strong potential to serve as a core component of data-driven marketing decision support systems that enhance strategic decision-making under uncertainty.

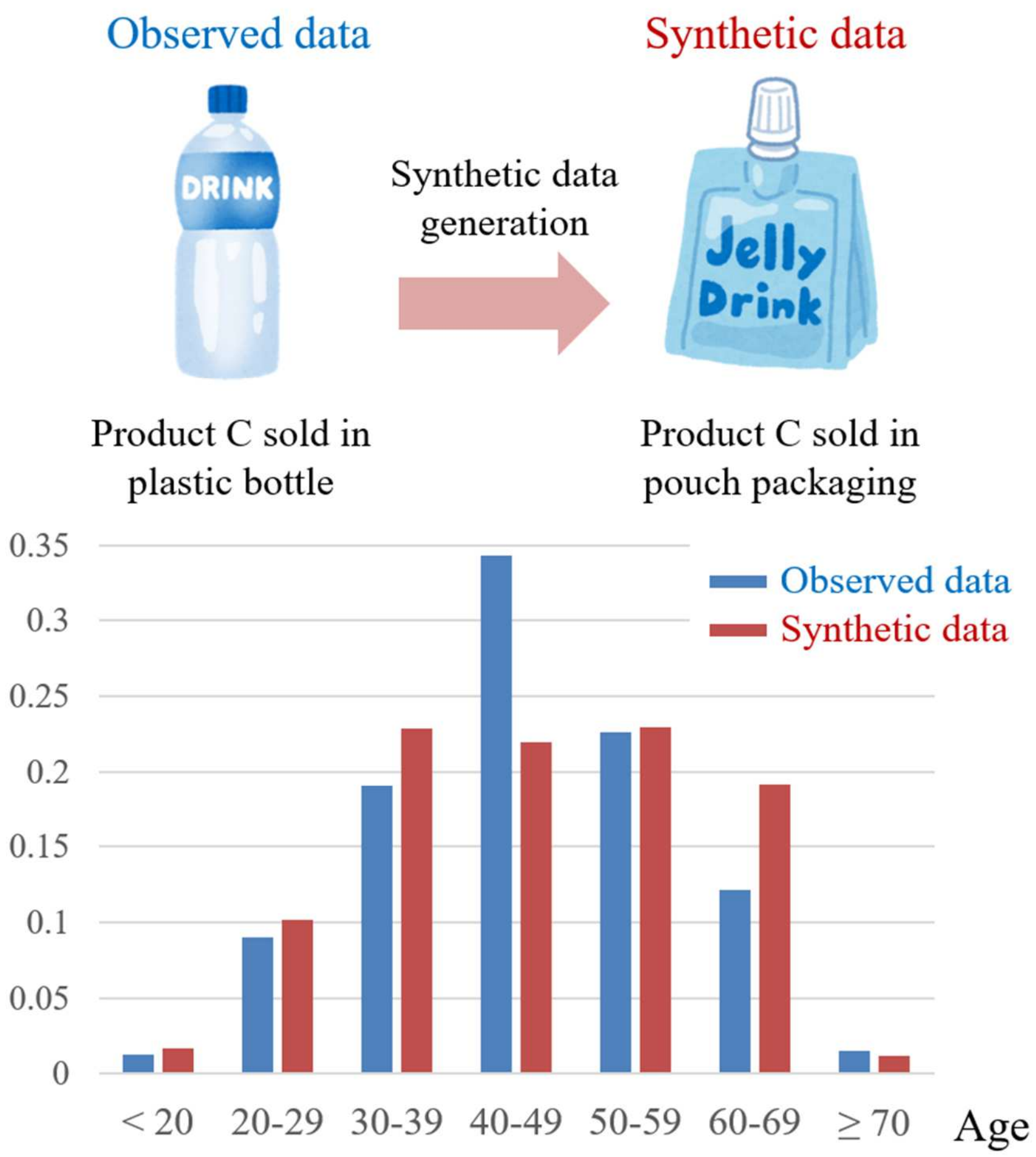

Figure 5. Simulation of line extension for container of product C

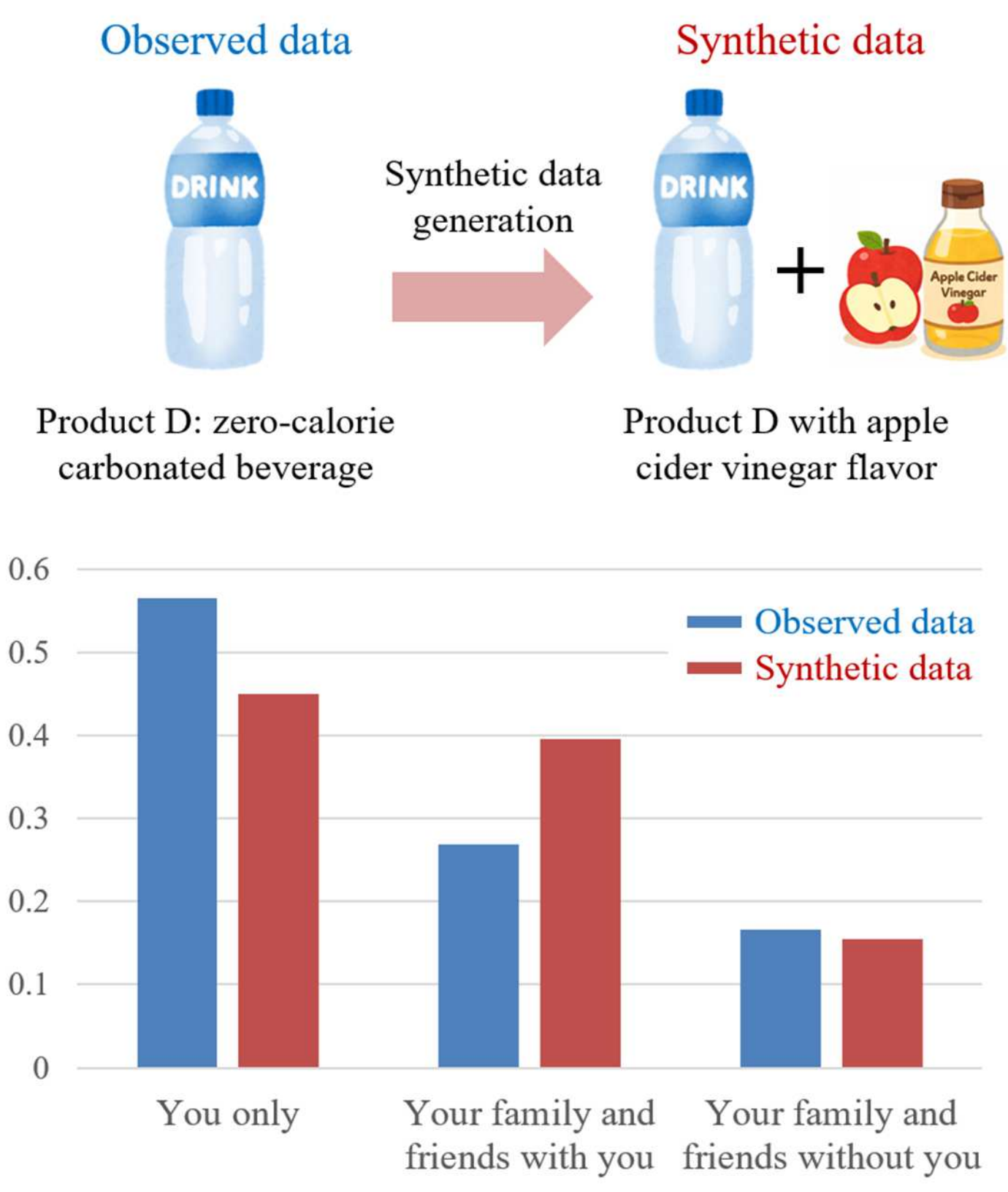

Figure 6. Simulation of line extension for adding apple cider vinegar of product D

**5.2 Contributions to knowledge-based systems research**

This study contributes to knowledge-based decision support research by reframing product line extension as a problem of data-driven knowledge generation and inference under uncertainty. Rather than treating product line extension solely as a conventional marketing decision, the study focuses on how explicit knowledge about consumer responses to product design changes can be acquired prior to market introduction. In practice, such knowledge has traditionally remained implicit, embedded in managerial experience or derived retrospectively from observed sales outcomes. The proposed approach advances decision support research by systematically transforming this implicit knowledge into explicit, data-driven representations that can be directly used in managerial reasoning.

The proposed CTVAE enables the generation of synthetic consumer attribute distributions conditioned on controllable product design variables. From a decision support perspective, this capability extends beyond conventional predictive models that are limited to supervised estimation on observed data. By generating plausible consumer profiles for hypothetical and as-yet-unobserved products, the CTVAE supports knowledge inference in decision contexts where empirical evidence is inherently unavailable at the time of decision-making.

Moreover, the generative and conditional structure of the proposed framework facilitates systematic what-if analysis, allowing decision-makers to explore alternative product design configurations and assess their potential impact on consumer segments and usage contexts. In this respect, the proposed model functions as a knowledge-based decision support mechanism that enables reasoning over counterfactual scenarios rather than merely providing point predictions or descriptive analytics. These characteristics position the CTVAE as a meaningful contribution to the decision support systems literature, particularly in advancing the role of deep generative models as core components of knowledge-driven systems for strategic decision-making under uncertainty.

### 5.3 Advantages of conditional generative modeling on tabular data

Unlike image or text domains, data used in marketing analytics and decision support systems are predominantly represented as structured tabular knowledge composed of heterogeneous numerical and categorical attributes. The experimental results demonstrate that the proposed CTVAE effectively captures complex dependency structures between product design attributes and consumer characteristics within such tabular representations. From a decision support perspective, this capability is critical, as it enables explicit modeling of how controllable design variables relate to downstream consumer outcomes.

A key advantage of the proposed framework lies in its conditional generative structure, which allows product attributes to function as explicit control variables in the knowledge generation and inference process. This design enables decision-makers to reason under hypothetical conditions by systematically manipulating product attributes and observing the resulting changes in inferred consumer attribute distributions. Rather than generating unconditional synthetic samples, the CTVAE supports structured exploration of alternative decision scenarios, which is a fundamental requirement in decision support systems aimed at strategic planning and pre-launch evaluation.

Compared with existing baseline models, the superior performance of the CTVAE suggests that conditional generative modeling is particularly well suited for knowledge acquisition and inference in structured decision-making environments. In such environments, the objective extends beyond predictive accuracy to include scenario analysis, comparative evaluation of alternatives, and explanatory reasoning about potential outcomes. The proposed model aligns closely with these requirements by enabling controlled, interpretable generation of synthetic decision-relevant knowledge.

From a computational perspective, existing tabular generative models such as CTGAN and TVAE often incur substantial overhead when performing conditional sampling with multiple control variables. This limitation stems from the need to implicitly model complex interactions between conditional inputs and latent representations, resulting in increased model complexity and optimization costs as the dimensionality of the conditional space grows. In contrast, the proposed

CTVAE explicitly incorporates conditional probabilities into the generative framework, thereby mitigating these scalability issues. Consequently, from both computational and representational viewpoints, the CTVAE provides a more efficient and scalable solution for synthetic knowledge generation in decision support applications involving product line extension and design exploration.

### 5.4 Performance evaluation

Based on the results presented in Section 4, the proposed CTVAE demonstrates strong potential as a generative inference model for knowledge-driven decision support in product line extension scenarios. Through holdout sample validation, the CTVAE consistently outperformed existing tabular generative models, including TVAE and CTGAN, in terms of predictive accuracy. These results indicate that the proposed model acquires higher-quality internal representations of the relationships between product attributes and consumer characteristics, which form the basis for reliable inference under hypothetical conditions. Beyond quantitative performance metrics, the case-based validations presented in Section 4.3 show that the proposed framework can generate concrete and plausible hypothetical scenarios that are directly meaningful for managerial decision-making. Such scenario-based outputs are particularly valuable in decision support systems, where the primary objective is not only accurate estimation but also the facilitation of reasoning under uncertainty. The ability to generate interpretable synthetic examples enhances the practical utility of the model as a mechanism for supporting exploratory analysis and informed judgment.

Nevertheless, further investigation is required to identify the conditions under which the proposed model produces reliable and actionable decision knowledge, as well as those under which the generated outputs may become less informative or operationally impractical. Clarifying these boundary conditions is essential for establishing robust evaluation criteria tailored to generative decision support models. Future research should therefore focus on developing systematic evaluation frameworks that assess not only predictive accuracy, but also the validity, stability, and decision relevance of generated knowledge. Such evaluation methodologies would contribute to advancing synthetic data generation research within the broader decision support systems literature, particularly for modeling complex and uncertain consumer behavior.

## 6. Conclusion

This study presented a novel deep generative framework designed to support decision-making in product line extension scenarios by acquiring and inferring consumer knowledge from historical purchase data. By explicitly modeling the joint distribution of product attributes and consumer characteristics, the proposed CTVAE enables the generation of synthetic consumer attribute distributions conditioned on hypothetical product designs. This capability allows decision-makers to conduct systematic what-if analyses prior to market introduction, thereby addressing a fundamental challenge in strategic product planning under uncertainty.

From a decision support systems perspective, the proposed approach transforms implicit, experience-based marketing knowledge into explicit, data-driven representations that can be directly used for scenario evaluation and comparative analysis. Experimental results demonstrated that the CTVAE outperforms existing tabular data generation methods in predicting consumer attribute changes. More importantly, beyond predictive accuracy, the generated synthetic knowledge supports simulation-based reasoning about alternative line extension strategies. This enables quantitative assessment of potential cannibalization risks, identification of shifts in target consumer segments, and exploration of product attribute configurations that align with anticipated consumption contexts. These characteristics highlight the effectiveness of deep generative models as core components of knowledge-driven decision support systems.

Despite these contributions, several limitations remain. The current framework assumes static consumer attributes and does not explicitly model temporal dynamics in consumer behavior or learning effects over time. In addition, the analysis is based solely on structured tabular data and does not incorporate external or unstructured information sources, such as textual product reviews, social media content, or expert knowledge. Addressing these limitations represents an important avenue for future research. Future work may extend the proposed framework by integrating temporal generative models, incorporating heterogeneous knowledge sources through knowledge graphs, and embedding explainable AI mechanisms to improve transparency and user trust in the inferred results. Such extensions would further enhance the applicability of generative modeling within decision support systems, contributing to the development of explainable, adaptive, and scalable decision support environments for complex marketing and product design decisions.

Acknowledgment

This work was supported by JSPS KAKENHI Grant Numbers JP23K25556, JP24K16458.

**References**

Aaker, D.A., Keller, K.L., 1990. Consumer evaluations of brand extensions. J. Mark. 54, 27–41. https://doi.org/10.1177/002224299005400102.

Borisov, V., Leemann, T., Seßler, K., Haug, J., Pawelczyk, M., Kasneci, G., 2024. Deep neural networks and tabular data: A survey, in: IEEE Trans. Neural Netw. Learn. Syst. 35, 7499–7519. https://doi.org/10.1109/TNNLS.2022.3229161.

Boush, D.M., Loken, B., 1991. A process-tracing study of brand extension evaluation. J. Mark. Res. 28, 16–28. https://doi.org/10.1177/002224379102800102.

Carlo, M., Ferilli, G., d'Angella, F., Buscema, M., 2021. Artificial intelligence to design collaborative strategy: An application to urban destinations. J. Bus. Res. 129, 936–948.

https://doi.org/10.1016/j.jbusres.2020.09.013

Chan, H., Choi, T., 2025. Using generative artificial intelligence (GenAI) in marketing: Development and practices. J. Bus. Res. 191, 115276. https://doi.org/10.1016/j.jbusres.2025.115276

Choi, E., Biswal, S., Malin, B., Duke, J., Stewart, W.F., Sun, J., 2017. Generating multi-label discrete patient records using generative adversarial networks, in: Proc. 2nd Mach. Learn. Healthcare Conf., pp. 286–305.

A. Mottini, A. Lheritier, and R. Acuna-Agost, "Airline passenger name record generation using generative adversarial networks," 2018, arXiv:1807.06657.

Clark Sinapuelas, I., Ram Sisodiya, S., 2010. Do line extensions influence parent brand equity? An investigation of supermarket packaged goods. J. Prod. Brand Manag. 19, 18–26. https://doi.org/10.1108/10610421011018356.

Crawford, M., Di Benedetto, A., 2010. New Products Management, tenth ed. Irwin Professional Publishing.

Darabi, S., Elor, Y., 2021. Synthesising Multi-modal Minority Samples for Tabular Data. arXiv:2105.08204. https://doi.org/10.1109/ACCESS.2021.3116481.

Fan, J., Chen, J., Liu, T., Shen, Y., Li, G., Du, X., 2020. Relational data synthesis using generative adversarial networks: A design space exploration. Proc. VLDB Endow. 13, 1962–1975. https://doi.org/10.14778/3407790.3407802.

Fonseca, J., Bacao, F., 2023. Tabular and latent space synthetic data generation: A literature review. J. Big Data. 10, 115. https://doi.org/10.1186/s40537-023-00792-7.

Gm, H., Gourisaria, M.K., Pandey, M., Rautaray, S.S., 2020. A comprehensive survey and analysis of generative models in machine learning. Comput. Sci. Rev. 38, 100285. https://doi.org/10.1016/j.cosrev.2020.100285.

He, C., Ke, S., Zhang, X., 2022. A model of product line marketing. Manag. Sci. 68, 6100–6115. https://doi.org/10.1287/mnsc.2021.4193.

Huiru, W., Jinhui, S., Jianying, F., Huiru, F., Zhijian, Z., Weisong, M., 2018. An agent-based modeling and simulation of consumers' purchase behavior for wine consumption. IFAC PapersOnLine. 51, 843–848. https://doi.org/10.1016/j.ifacol.2018.08.089.

Johann, T.I., Otte, K., Prasser, F., Dieterich, C., 2025, anonymize or synthesize? – Privacy-preserving methods for heart failure score analytics, Eur. Heart J. Digit. Health. 6, 147–154. https://doi.org/10.1093/ehjdh/ztae083

Keller, K.L., 2008. Strategic Brand Management: Building, Measuring, and Managing Brand Equity, third ed. Prentice Hall.

Kingma, D.P., Rezende, D.J., Mohamed, S., Welling, M. Semi-supervised learning with deep generative models, in: Proc. 27th International Conference on Neural Information Processing Systems, pp. 3581–3589.

Ładyżyński, P., Żbikowski, K., Gawrysiak, P., 2019. Direct marketing campaigns in retail banking with the use of deep learning and random forests. Expert Syst. Appl. 134, 28–35. https://doi.org/10.1016/j.eswa.2019.05.020.

Liao, S.H., Chen, C.M., Wu, C.H., 2008. Mining customer knowledge for product line and brand extension in retailing. Expert Syst. Appl. 34, 1763–1776. https://doi.org/10.1016/j.eswa.2007.01.036.

Liu, H., Yao, X., Zhang, L., 2024. Unveiling consumer preferences: A two-stage deep learning approach to enhance accuracy in multi-channel retail sales forecasting. Expert Syst. Appl, 257, 125066. https://doi.org/10.1016/j.eswa.2024.125066

Ma, C., Tschiatschek, S., Turner, R., Hernández-Lobato, J.M., Zhang, C., 2020. 'VAEM: A deep generative model for heterogeneous mixed type data,' in Proc. Adv. Neural Inf. Process. Syst. 33, 1–11.

Malhotra, N., 2019. Marketing Research: An Applied Orientation, seventh ed. Pearson Education.

Mamta, K., Sangwan, S., 2024. AaPiDL: An ensemble deep learning-based predictive framework for analyzing customer behaviour and enhancing sales in e-commerce systems. Int. J. Inf. Tecnol. 16, 3019–3025. https://doi.org/10.1007/s41870-024-01796-z.

Mirashk, H., Albadvi, A., Kargari, M., Javide, M., Eshghi, A., Shahidi, G., 2019. Using RNN to predict customer behavior in high volume transactional data, in: Grandinetti, L., Mirtaheri, S., Shahbazian, R. (Eds.) High-Performance Computing and Big Data Analysis. Commun. Comput. Inf. Sci. TopHPC 2019, vol 891. Springer, Cham. https://doi.org/10.1007/978-3-030-33495-6_30.

Nikolentzos, G., Vazirgiannis, M., Xypolopoulos, C., Lingman, M., Brandt, E.G., 2023. Synthetic electronic health records generated with variational graph autoencoders. npj Digit. Med. 6, 83. https://doi.org/10.1038/s41746-023-00822-x.

Pandey, G., Dukkipati, A., 2017. Variational methods for conditional multimodal deep learning, in: Proc. 2017 International Joint Conference on Neural Networks (IJCNN), Anchorage, AK, USA. IEEE, New York, pp. 308–315. https://doi.org/10.1109/IJCNN.2017.7965870.

Park, N., Mohammadi, M., Gorde, K., Jajodia, S., Park, H., Kim, Y., 2018. Data synthesis based on generative adversarial networks. Proc. VLDB Endow. 11, 1071–1083. https://doi.org/10.14778/3231751.3231757.

Park, S.K., Sela, A., 2020. Product lineups: The more you search, the less you find. J. Con. Res. 47, 40–55. https://doi.org/10.1093/jcr/ucaa001.

Vrinda Kadiyali, Naufel Vilcassim, Pradeep Chintagunta, Product line extensions and competitive market interactions: An empirical analysis, Journal of Econometrics, Volume 89, Issues 1–2, Pages 339-363, 1998

Rand, W., Rust, R.T., 2011. Agent-based modeling in marketing: Guidelines for rigor. Int. J. Res. Mark. 28, 181–193. https://doi.org/10.1016/j.ijresmar.2011.04.002.

Reddy, S.K., Holak, S.L., Bhat, S., 1994. To extend or not to extend: Success determinants of line extensions. J. Mark. Res. 31, 243–262. https://doi.org/10.1177/002224379403100208.

Sahakyan, M., Aung, Z., Rahwan, T., 2021. Explainable artificial intelligence for tabular data: A survey, in: IEEE Access. 9, 135392–135422. https://doi.org/10.1109/ACCESS.2021.3116481.

SDMetrics DataCebo (a). TVComplement. Accessed Novenver, 2024, vol 25. https://docs.sdv.dev/sdmetrics/metrics/metrics-glossary/tvcomplement.

SDMetrics DataCebo (b). KSComplement. Accessed Novenver, 2024, vol 25. https://docs.sdv.dev/sdmetrics/metrics/metrics-glossary/kscomplement.

Sohn, K., Yan, X., Lee, H., 2015. Learning structured output representation using deep conditional generative models, in: Proc. 28th International Conference on Neural Information Processing Systems, Montreal, QC, Canada, pp. 3483–3491.

Sun, Q., Feng, X., Zhao, S., Cao, H., Li, S., Yao, Y., 2021. Deep learning based customer preferences analysis in Industry 4.0 environment. Mob. Netw. Appl. 26, 2329–2340. https://doi.org/10.1007/s11036-021-01830-5.

Tang, X., Yan, J., Li, Y., 2023. Supervised multi-layer conditional variational auto-encoder for process modeling and soft sensor. Sensors (Basel). 23, 9175. https://doi.org/10.3390/s23229175.

Titar, R.R., Ramanathan, M., 2024. Variational autoencoders for generative modeling of drug dosing determinants in renal, hepatic, metabolic, and cardiac disease states. Clin. Transl. Sci. 17, e13872. https://doi.org/10.1111/cts.13872.

Xu, L., Skoularidou, M., Cuesta-Infante, A., Veeramachanen, K., 2019. Modeling Tabular data using Conditional GAN, in: Proc. NeulIPS, pp. 7335–7345.

Zhang, J., Cormode, G., Procopiuc, C.M., Srivastava, D., Xiao, X., 2017. PrivBayes: Private data release via Bayesian networks. ACM Trans. Database Syst. 42, 1–41, Article No.: 25. https://doi.org/10.1145/3134428.

Zhang, T., Zhang, D., 2007. Agent-based simulation of consumer purchase decision-making and the decoy effect. J. Bus. Res. 60, 912–922. https://doi.org/10.1016/j.jbusres.2007.02.006.

Zhang, W., Feng, J., Li, F., 2024. Deep learning-based customer lifetime value prediction in imbalanced data scenarios: A case study, in: Tan, Y., Shi, Y. (Eds.) Advances in Swarm Intelligence. Lect. Notes Comput. Sci. ICSI 2024, vol 14789. Springer, Singapore. https://doi.org/10.1007/978-981-97-7184-4_18.

Zhu, Y., He, W., Huang, M., 2023. Influence prediction model for marketing campaigns on e-commerce platforms. Expert Syst. Appl, 211, 118575. https://doi.org/10.1016/j.eswa.2022.118575.